\documentclass[twoside,11pt]{article}

\usepackage{blindtext}
\usepackage{xcolor}         
\usepackage{graphicx}
\usepackage{mdframed}  
\usepackage{subcaption}  
\usepackage{enumitem}
\usepackage{wrapfig}  
\usepackage[abbrvbib, preprint]{jmlr2e}
\usepackage{booktabs}  
\usepackage{hyperref}
\usepackage{xcolor}  
\definecolor{myColor}{RGB}{240,190,138}  
\usepackage{caption}  
\hypersetup{  
  colorlinks=true,  
  citecolor=myColor  
}  

\usepackage{lastpage}

\firstpageno{1}

\begin{document}

\title{Introspective Tips: Large Language Model for \\In-Context Decision Making}

\author{Liting Chen$^1$, Lu Wang$^1$, Hang Dong$^1$, Yali Du$^2$, Jie Yan$^1$, Fangkai Yang$^1$, Shuang Li$^3$, Pu Zhao$^1$, Si Qin$^1$, Saravan Rajmohan$^1$, Qingwei Lin$^1$, \\Dongmei Zhang$^1$}  
\newcommand{\authorclean}{Liting Chen, Lu Wang, Hang Dong, Yali Du, Jie Yan, Fangkai Yang, Shuang Li, Pu Zhao, Si Qin, Saravan Rajmohan, Qingwei Lin, Dongmei Zhang}  
  

\editor{My editor}

\maketitle
\begin{center}  
  \textsuperscript{1}Microsoft
  \textsuperscript{2} Kings College London
  \textsuperscript{3} The Chinese University of Hong Kong (Shenzhen)
\end{center}  

\begin{abstract}

The emergence of large language models (LLMs) has substantially influenced natural language processing, demonstrating exceptional results across various tasks. In this study, we employ ``Introspective Tips" to facilitate LLMs in self-optimizing their decision-making. By introspectively examining trajectories, LLM refines its policy by generating succinct and valuable tips.  Our method enhances the agent's performance in both few-shot and zero-shot learning situations by considering three essential scenarios: learning from the agent's past experiences, integrating expert demonstrations, and generalizing across diverse games. Importantly, we accomplish these improvements without fine-tuning the LLM parameters; rather, we adjust the prompt to generalize insights from the three aforementioned situations. Our framework not only supports but also emphasizes the advantage of employing LLM in in-contxt decision-making. Experiments involving over 100 games in TextWorld illustrate the superior performance of our approach.
\end{abstract}

\section{Introduction}




Large Language Models (LLMs), including OpenAI's GPT-3.5~\citep{ouyang2022training}, GPT-4~\citep{OpenAI2023GPT4TR}, Google's PaLM~\citep{chowdhery2022palm} and other benchmarks~\citep{meta2023introducing, alpaca} have consistently achieved remarkable performance across various NLP tasks. The integration of LLMs into decision-making tasks~\citep{huang2022language,ahn2022can,kwon2023reward,brooks2022context,yao2022react} has garnered significant attention, as it presents an opportunity to develop decision-making agents that can emulate human-like cognitive processes~\citep{shevlin2019limits}.
In decision-making tasks, such as in the domain of Reinforcement Learning (RL), limited interactions hinder optimal policy learning~\citep{yarats2021improving}. 
Owing to the power of LLMs, which have undergone extensive pre-training on vast amounts of data, agents can leverage the generalization capabilities of LLMs to enhance their performance across different tasks. In addition, common sense knowledge is a valuable asset in safety decision-making tasks~\citep{brunke2022safe}, and LLMs possess a wealth of this knowledge due to their extensive training and tuned with human feedback~\citep{ouyang2022training}. 
By leveraging the inherent common sense knowledge within LLMs, decision-making agents can make more informed decisions, effectively tackling challenges such as sparse rewards and enhancing the learning process. 

However, LLMs sometimes demonstrate errors or hallucinations~\citep{ji2023survey,peng2023check}, especially in domain-specific scenarios. Recent works focus on designing self-correction mechanisms to enhance the LLM's decision-making performance. For example, Chain-of-thought (CoT)~\citep{wei2022chain} concentrates on static reasoning for one-step actions without self-correction. ReAct~\citep{yao2022react} and Reflexion~\citep{shinn2023reflexion} focus on dynamic reasoning based on historical interactions to enable better decision-making performance. However, these works primarily focus on generating instructions or reflections specific to individual agents, hindering the adoption of the generalization and universality abilities of LLMs. Furthermore, the absence of correction for the generated reflections can lead to errors during the decision-making process.

\begin{figure}[htb] 
    \centering
    \includegraphics[width=1\linewidth]{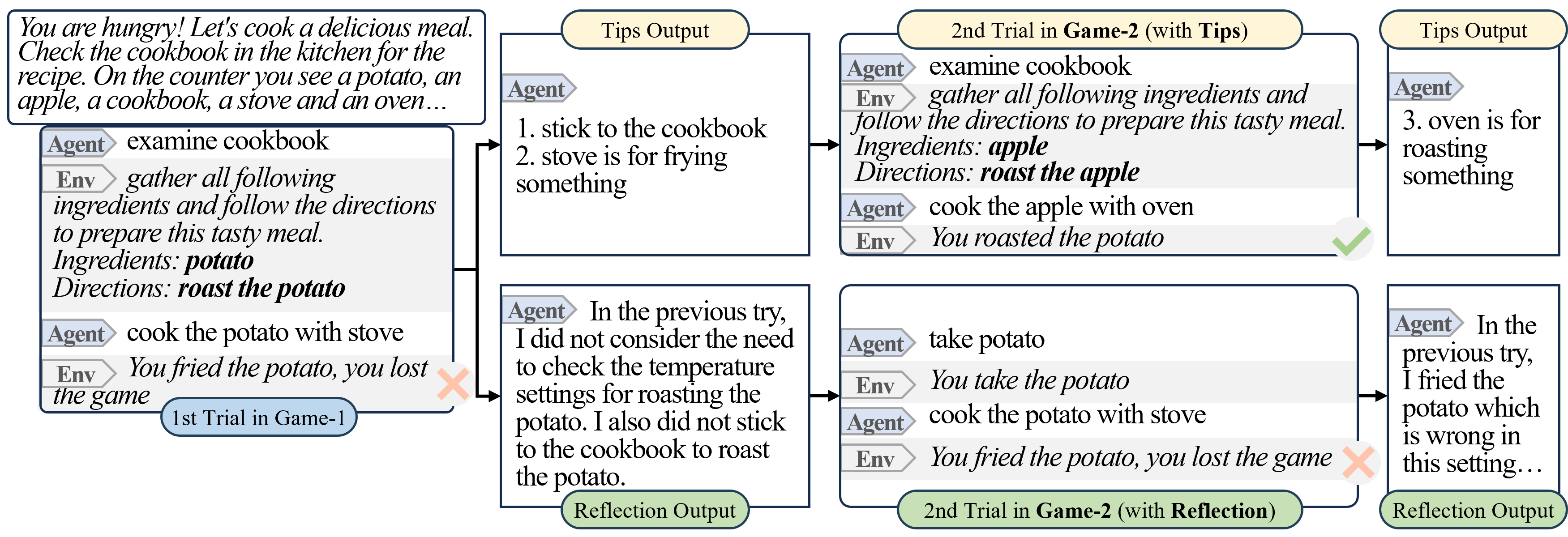}
    \caption{In both Game 1 and Game 2, the agent is tasked with checking the cookbook and cooking a meal. The cookbooks for each game are different, but the initial state remains the same. After playing in Game 1, the agent is asked to participate in Game 2, using tips or reflections generated from the Game 1 trajectory. The agent's reflection focuses on the errors made in Game 1 and is specific to that game. Tips, on the other hand, offer more general and high-level information about the game. Utilizing tips from Game 1, the agent learns that the stove is for frying ingredients and uses the oven to roast the apple. The agent subsequently updates its knowledge about the game by incorporating a third tip. However, when using reflections, the agent receives an irrelevant message about potatoes, which leads to its failure to check the cookbook and complete the task in Game 2. In this case, the general tips prove to be more helpful in guiding the agent's behavior and improving its performance, whereas the specific reflections from Game 1 do not provide useful information for Game 2. }
    \label{fig:intro_example}
\end{figure}

Humans possess high generalization and self-evolution capabilities due to their natural ability to summarize \textit{tips} through introspection~\citep{cox1996introspective,van2005cognitive}. Introspection enables humans to reflect on their past experiences (successes or failures), distill key insights, and extract valuable lessons that can be applied to new situations. 
Inspired by the remarkable introspection capabilities of humans, we introduce ``Introspective Tips'' (or ``Tips'' for short) as a novel prompt-based paradigm, empowering an LLM decision-making agent with self-optimizing capabilities via learning tips from trajectories of itself, experts, and other environments. 
Figure~\ref{fig:intro_example} provides an example of introspective tips in a text-based game. 

Distinct from reflection~\citep{ shinn2023reflexion}, which is a self-analysis process that delves into an agent's actions and experiences in detail, "tips" are compact pieces of information offering high-level guidance or suggestions. The goal of tips is to provide succinct and practical advice for enhancing decision-making performance without requiring an in-depth examination of past trajectories. In contrast, reflections can only be derived from the agent itself and primarily concentrate on learning from failures. Tips, on the other hand, can be acquired from the trajectories of other agents and human experts, focusing on learning from both successes and failures. For example, in the game setting, the agent can generalize tips to correctly use oven based on its successful trajectory, but reflection may focus more on failures.

Additionally, in contrast to using original trajectories as prompts, introspective tips serve as condensed and comprehensive insights extracted from lengthy trajectories. This allows LLMs to overcome the challenges in discerning the intricate relationship between dependent actions and sparse rewards across multiple trajectories~\citep{adhikari2020learning}. Introspective tips provide distinct advantages in terms of their applicability and generalization across different agents and contexts, as opposed to previous prompting methods such as reflections. Unlike reflections that target specific agents, tips can be shared among different agents, allowing for higher generalization performance. In the game setting of Figure~\ref{fig:intro_example}, the tips generalized by the agent can be applied to various game settings with different recipes. In contrast, reflections are more specific to a single past trajectory.

Furthermore, designing suitable prompts for LLMs to generate introspective tips is a crucial challenge; manually crafting prompts can be burdensome. Therefore we propose a framework that dynamically adjusts the prompt based on insights derived from past trajectories or expert demonstrations through introspection.

Our contributions can be summarized as follows: 
\begin{itemize}[leftmargin=*]
    \item \textbf{Introspective Tips for Self-Optimizing Decision-Making:} We introduce Introspective Tips as a novel prompt-based paradigm, empowering LLM decision-making agents with self-optimizing capabilities for both few-shot and zero-shot scenarios. Unlike previous RL agents, Introspective Tips offers an end-to-end solution without the need for training or fine-tuning. It generates meaningful and easy-to-understand tips, facilitating human understanding and intervention in the decision-making process.
    \item \textbf{Multiple Strategies in different scenarios for Learning Introspective Tips:} We present distinct strategies applicable in different scenarios to prompt LLMs to learn Introspective Tips from their own trajectories, expert demonstrations, and multi-environment trajectories, which also shows the \textbf{generalization} and \textbf{adaptability} of Introspective Tips. 
    \item \textbf{A Dynamic Prompt Adjustment Framework:} To simplify the prompting engineering process, we introduce a framework that dynamically adjusts the prompt by leveraging insights from past trajectories or expert demonstrations through introspection. This framework streamlines the improvement process, enhances the model's adaptability, and enables more efficient decision-making.
    \item \textbf{Comprehensive Evaluation and Comparison:} Our extensive experiments encompass over 100 games in TextWorld~\citep{adhikari2020learning, cote2019textworld}, testing few-shot and  zero-shot learning scenarios. We evaluate the performance of our decision-making agent against state-of-the-art methods in the field~\citep{tuli2022learning}, highlighting the effectiveness and superiority of Introspective Tips. With tips generated from 48 trajectories, LLM agent performs better than previous deep learning method training for 100, 000 episodes in the unseen game with the highest difficulty level.
\end{itemize}

\section{Related work}
\textbf{Language model for decision making}
LLMs~\citep{OpenAI2023GPT4TR,chowdhery2022palm} have exhibited impressive proficiencies, facilitating their use in tasks beyond mere language generation and increasingly serving as policy models for decision-making in interactive settings~\citep{yang2023foundation}. \cite{wei2022chain} demonstrates that incorporating a series of intermediate reasoning steps can enhance decision-making abilities. \cite{yao2022react} introduces ReAct, a method for interleaved reasoning and action generation that fosters improved synergy between language comprehension and interactive decision-making tasks. \cite{shinn2023reflexion} presents Reflexion, a technique that equips an LLM-based agent with a self-reflective LLM and a straightforward heuristic for detecting hallucination and inefficient action execution to examine the hypothesis. \cite{madaan2023self} adopts a similar strategy, enabling an LLM to offer feedback on its previously generated text and refine it to meet specific requirements.
When regarded as dialogue agents, LLMs can also be trained to learn from human feedback and optimize their output ~\citep{ouyang2022training,bai2022training}. With further training, \cite{li2022pre} constructs a general framework for decision-making tasks using pre-trained LMs, even in scenarios where language is neither provided as input nor output.
Other studies~\citep{singh2022progprompt,huang2022language,huang2022inner,liang2022code,vemprala2023chatgpt} have explored innovative strategies involving prompt engineering and the utilization of high-level function libraries to enhance the capabilities of LLMs. 


Recent attempts explored different aspects of LLMs for decision-making. \cite{huang2022language} and ~\cite{ahn2022can} use LLMs to generate plans or sub-goals that guide low-level Reinforcement Learning (RL) agents in taking actions.
\cite{kwon2023reward} utilize LLMs as proxy reward functions by prompting them with desired behaviors.
\cite{yao2022react} focus on enabling LLM agents to select actions in text-based environments. 
In addition, a recent approach considers LLMs as world models~\citep{brooks2022context}, where the agent learns the policy by interacting with the LLM-based world model.
In this paper, we focus on directly grounding LLM in decision-making to take action because the other three methods required learning an extra decision-making agent, which required more samples.

\textbf{In-context reinforcement learning}
In-context learning pertains to the ability of sequence prediction models to adapt to novel downstream tasks solely through the use of prompts without retraining or fine-tuning~\citep{lu2021pretrained, brown2020language,min2022rethinking}. When applied to reinforcement learning, in-context learning models can generalize to diverse downstream tasks when provided with contexts such as demonstrations and task information. \cite{laskin2022context} recasts RL as an across-episode sequential prediction problem, and trains a causal transformer to autoregressively predict actions based on preceding learning histories as context. \cite{brooks2022context} employs the LLM as a world model for planning future trajectories and executing decisions in-context. \cite{team2023human} develops a versatile in-context learning algorithm capable of adapting to new and open-ended 3D challenging problems as rapidly as humans, by training an RL agent at a large scale. \cite{lu2023structured} proposes to meta-learn across random linear projections of the observation and action spaces of randomly sampled DMControl tasks ~\citep{tassa2018deepmind}. Trained on an extensive dataset, Gato ~\citep{reed2022generalist} can generalize to new tasks by conditioning on demonstrations of the desired behavior.

\textbf{Text-based game}
Text-based games are typically turn-based experiences played via a command line terminal. During each turn, the game state is conveyed through multiple lines of text, which enables players to input text commands that modify the state according to their preferences ~\citep{liu2022learning,hendrycks2021would,osborne2022survey}. Text-based games can be formally characterized as partially observable Markov decision processes (POMDPs) ~\citep{cote2019textworld}, considering that the agent only observes partial information about the environment at each turn. Intrinsic obstacles such as long-term dependencies, partial observation of current states, and sparse rewards, and complex action combinations render these games particularly challenging. Various deep learning approaches have been employed to address text-based games ~\citep{xu2022perceiving,yin2019comprehensible, ammanabrolu2020graph,kimura2021neuro}. Focusing on tasks in the TextWorld domain ~\citep{cote2019textworld}, \cite{adhikari2020learning} explores learning graph-structured state representations via a data-driven approach, introducing the Graph Aided Transformer Agent (GATA) that learns to construct and update graph-structured beliefs while optimizing rewards. Building upon their work, \cite{tuli2022learning} equips GATA with an internal structured representation of natural language instructions using Linear Temporal Logic (LTL) to enhance the instruction-following capabilities of text-based game agents.






\section{Method}
\definecolor{trial1Color}{rgb}{0.980392, 0.858824, 0.776471}  
\definecolor{trial2Color}{rgb}{0.780392, 0.882353, 0.709804}  
\definecolor{trial3Color}{rgb}{1.0, 0.933333, 0.741176}  
\definecolor{trial4Color}{rgb}{0.909804, 0.85098, 0.952941}  
In this section, we elaborate on the method employed to leverage the capabilities of LLMs in sequential decision-making tasks, particularly text-based games, by addressing LLMs inherent limitations and capitalizing on their strengths. LLMs' proficiency in understanding and generating human-like text renders them promising candidates for tasks involving natural language processing, such as text-based games. However, they lack specific domain knowledge when asked to make decisions in certain tasks thus may not reach the greatest potential when used directly. 

We conjecture that the function space of generative LLMs $P(\theta)$ is sufficiently expansive to encompass the function of an expert policy in text-based games. The action space of policy $\pi$ is constrained  by text output. By selecting an appropriate prompt $pr$ containing enough domain knowledge, we can derive a model $P(\theta|pr)$ that can function as a $\pi$ in decision making, bridging the gap between LLMs and reinforcement learning. Through incorporating strategies including learning from past experiences, expert demonstrations, and multiple games, we aim to develop a versatile and robust framework that excels in a wide range of gaming scenarios. This approach ultimately opens up the journey of discovering the potential of LLMs in the domain of complex sequential decision-making tasks.  

\subsection{Challenges and Solutions for LLMs in Decision-Making Problems} 

There are several major challenges when applying Language Learning Models (LLMs) to solve decision-making problems, including self-optimization, input length limitations, and prompt dependence. In this section, we discuss these challenges and potential solutions.

\textbf{Self-Optimizing decision-making }
LLMs can exhibit errors or hallucinations, particularly in domain-specific scenarios. Developing self-optimizing mechanisms to enable LLMs to correct their mistakes is essential for improving their decision-making performance. Moreover, allowing LLMs to self-optimize across various scenarios poses a significant challenge. Current approaches, such as self-reflection, attempt to rectify errors by concentrating on the model's own mistakes. However, this method has limitations, including the risk of LLMs becoming misled by their own errors during reflection and lacking a comprehensive understanding of the task or environment. To address these challenges, we employ "Introspective Tips" to facilitate LLMs in self-optimizing their decision-making. We empower LLMs to learn from their own trajectories, expert demonstrations, and trajectories in different environments. By extracting concise and actionable advice from these trajectories, the agent gains a better understanding of the environment. Furthermore, by refining its own tips, the problem of hallucination is alleviated.

\textbf{Limited input length} Language models like GPT-4~\citep{OpenAI2023GPT4TR} and PaLM~\citep{chowdhery2022palm} often encounter input length limitations. In  RL, agents typically face lengthy and complex trajectories, as well as sparse rewards that offer feedback only after numerous steps or interactions. Due to the restricted input capacity and extended trajectories,  LLMs struggle to discern the relationship between actions and rewards, ultimately impeding the acquisition of the optimal policy. To address these issues and optimize the available input space, we use  tips that effectively condense the training dataset or critical information generally found in classical RL settings. By incorporating these summaries, LLMs can better understand the relationships within the data and identify essential patterns that might be otherwise obscured by the sheer length or complexity of the trajectories. Moreover, providing concise, relevant information enables LLMs to focus on the crucial aspects of the problem at hand, potentially leading to more accurate results. This strategy can be particularly beneficial in scenarios where the agent must learn from limited data or adapt to changing environments, as the distilled information can help guide the learning process more efficiently.
  
\textbf{Prompt dependence:} The LLM's performance on a specific problem heavily relies on the quality and order of prompts. In order to solve the sequential decision problem in text-based games, that is to produce more relevant and valuable responses in each round, it is crucial to craft prompts in decision making. 
However, creating prompts using human interaction can be time-consuming and tedious. To address this issue, our method introduces a framework that dynamically adjusts the prompt based on past trajectories through introspection, streamlining the process and improving the model's adaptability. While designing a specific tip for one game can be laborious, we also explore the possibility of generating a universal prompt that can transform the LLM into an efficient RL agent across various games and contexts. This general prompt would further facilitate the LLM's ability to adapt and perform effectively in a wide range of decision-making situations.
  

\begin{figure}[htb] 
    \centering
    \includegraphics[width=1\linewidth]{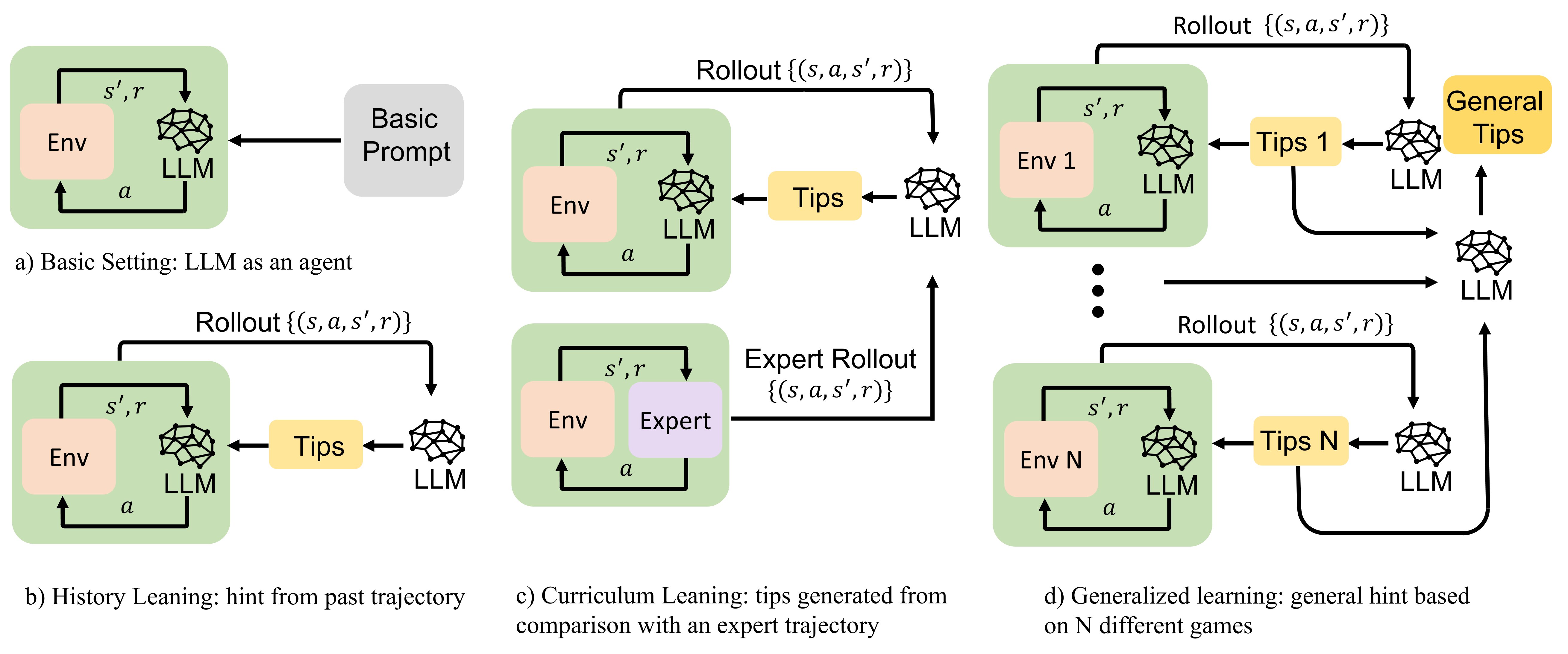}
    \caption{LLM as RL agent}
    \label{fig:side_by_side}
\end{figure}
  


  
\subsection{Basic setting}

In our basic setting (Fig 2a), we directly utilize an LLM as an agent participating in a text-based game. A sample interaction proceeds as follows:

\begin{wrapfigure}{r}{0.5\textwidth}
\vspace{-2em}
\begin{mdframed}[backgroundcolor=trial1Color, linewidth=0pt] 
\textbf{\textbf{system:}} You are an agent playing in a text-based game. All of your available actions are in the ActionList:\\
inventory() \# print player's inventory \\
... \textit{(see appendix for the full list of action)}
 Based on the game's description that I give you, provide me with only one action per step in the action list and wait for my response. (Following is the description of the first state in a TextWorld game.)
\end{mdframed}  
\begin{mdframed}[backgroundcolor=trial2Color, linewidth=0pt] 

\textbf{\textbf{Agent:}} inventory() 
\end{mdframed}  
\vspace{-2em}
\end{wrapfigure}
The system (user) clearly defines the role and action space for the LLM agent and provides an initial text-based description of the environment, setting the scene for the agent. The LLM-agent interprets the provided information and decides on an action based on its understanding of the game's context. The agent submits its chosen action as text, which the system interprets, processes, updates the environment, and provides new feedback, including information on invalid actions or a description of the updated state, to the agent. The LLM-agent continues to interact with the game, choosing actions and receiving feedback, until the game reaches its conclusion. The game concludes as a success if the agent completes all required steps, or as a failure if the agent takes erroneous actions or reaches the maximum number of turns. Throughout the entire process, the agent's goal is to navigate the game world and make decisions based on textual input and output.
  
\subsection{Strategies for three scenarios}

We have developed strategies for three scenarios to enhance the agent's learning addressing the two challenges:

\textbf{Learning from its own history:} The agent employs a history replay method to generate tips for self-improvement. The agent is prompted as the basic setting to finish a basic trial. If the agent fails, the agent creates several tips to address its past mistakes. In subsequent attempts, these tips serve as prompts. If the agent fails even if provided with tips, it is prompted to generate more effective tips for itself. The agent is also provided with all its previous actions that led to failure in the game. By introspectively analyzing its past actions and their outcomes, the agent generates valuable insights to refine its policy. This self-enhancement process enables the agent to overcome obstacles and make better choices in future gameplay. Moreover, since tips and incorrect actions are shorter than an entire trajectory, the agent can learn from a more extensive history than relying solely on past experiences as memory (Fig 2b).

\textbf{Learning from expert demonstrations:} The agent evaluates its performance by comparing it to expert demonstrations and makes adjustments as necessary. In this setting, the agent is instructed to observe both the actions leading the expert agent to achieve the goal and the actions causing it to fail the game. By examining successful strategies and tactics employed by experts, the agent can pinpoint areas for improvement and generat tips that can help modify its actions accordingly. This learning approach is faster compared to a purely trial-and-error method. For instance, the agent can learn the correct usage of an appliance by contrasting its failed attempts with the successful actions demonstrated by an expert in a text-based game, rather than experimenting with all possible actions.

The agent generates tips based on these observations, and these tips serve as new prompts when playing subsequent rounds. If the agent fails even when provided with tips, it is prompted to reflect on the given advice. This method enables knowledge transfer from expert demonstrations, resulting in a more efficient and informed decision-making process. Feedback from the environment acts as guidance for the LLM model, assisting it in determining the accuracy and effectiveness of its generated tips (Fig 2c).

\textbf{Learning from multiple games:} In this scenario, agents are allowed to participate in several games. All final tips (tips that lead to a successful trial) generated across these games are sent to another LLM, which is then tasked with generating general tips to help the agent become an expert across various game settings. In this context, the LLM leverages its knowledge of summarization to produce comprehensive and valuable tips that enable the agent to improve its performance in a wide range of games. These games share some similarities but with some information different. For example, in the cooking game, these games share a similar theme (cooking in a modern house), similar text commands, and similar entities (i.e., interactable objects within the games), but with different cookbooks and maps involved in the game. The LLM generalizes knowledge across diverse games, functioning similarly to a Meta-RL agent. The general tips are served as prompts in the unseen test game. This ability allows the LLM to perform effectively in a broad range of text-based games, positioning it as a powerful tool for reinforcement learning in complex environments (see Fig 2d).

By incorporating these strategies in the three scenarios, our method presents a robust framework for exploiting the potential of LLMs in reinforcement learning for text-based games. Learning from past experiences, expert demonstrations, and multiple games allows the agent to adapt and excel in various gaming scenarios, ultimately demonstrating the versatility and effectiveness of LLMs in the realm of reinforcement learning.

\section{Experimental Results}

\begin{figure}[htb] 
    \centering
    \includegraphics[scale=0.53]{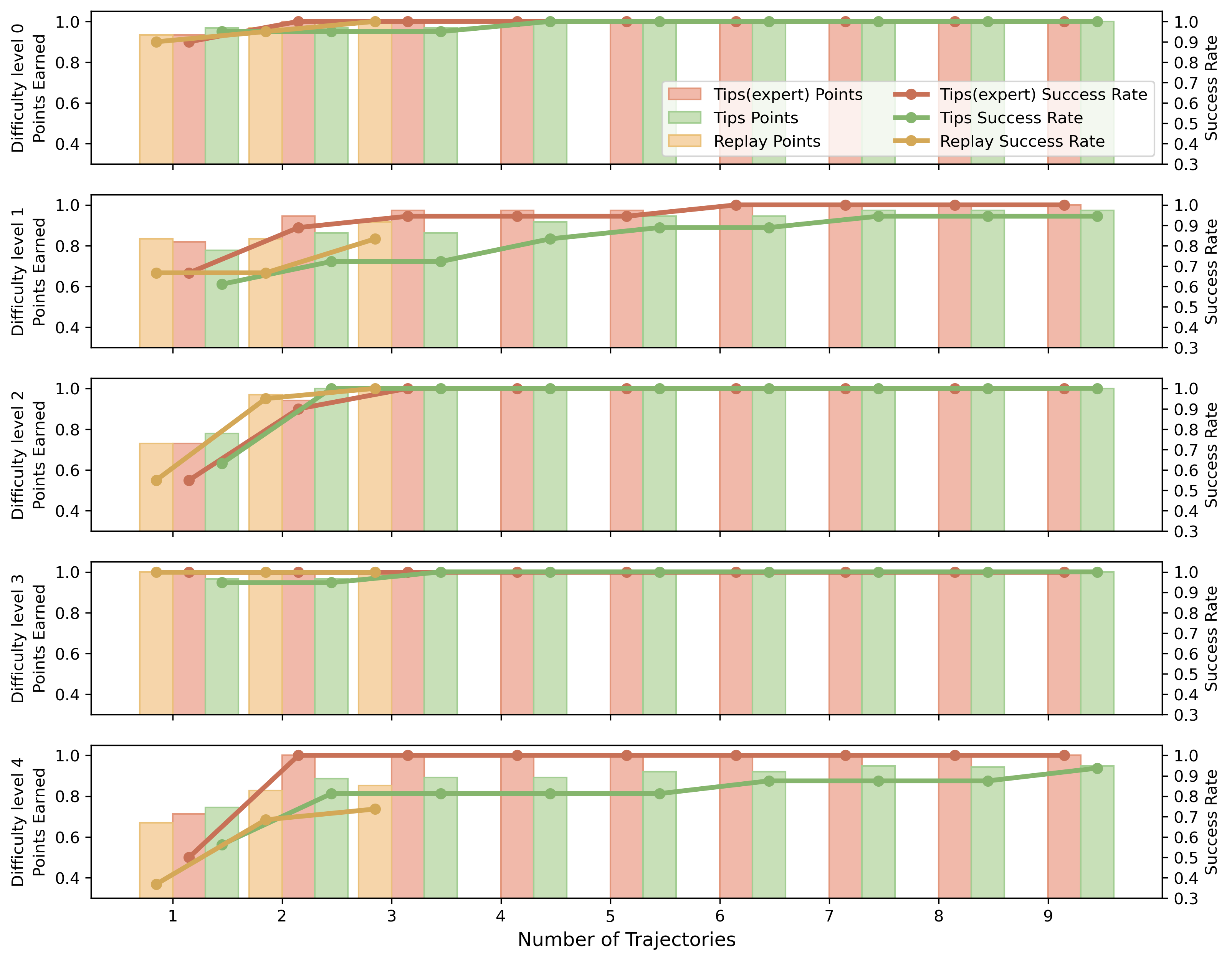}
    \caption{Few shot performance over different difficulty levels of games}
    \label{fig:2}
\end{figure}
We conduct experiments on text-based games to assess the performance of LLMs as agents and address three key questions: Q1: Can LLMs learn from their historical trajectories and improve their performance by reflecting on different tips? Q2: Can LLM learn from expert demonstrations and can expert demonstration boost the learning of different strategies? Q3: Can LLMs generalize some tips to play in various game settings?  For Q1 and Q2, we analyze the few-shot decision-making capabilities of LLMs, while for Q3, we concentrate on their zero-shot decision-making abilities.

\subsection {Experiment setup}
\begin{table}[ht]
\caption{Statistics of The Game}
\label{tb:1}
\footnotesize  
\centering

\begin{tabular}{cccccc}  
\toprule  
Level & \#Ingredients & \#Locations & Points& (Cook, Cut, Open )\\  
\midrule  
0 & 1 & 1 & 3  & (\texttimes, \texttimes, \checkmark) \\  
1 & 1 & 1 & 4  & (\texttimes, \checkmark, \checkmark) \\  
2 & 1 & 1 & 5  & (\checkmark, \checkmark, \checkmark)  \\  
3 & 1 & 9 & 3  & (\texttimes, \texttimes, \checkmark)\\  
4 & 3 & 6 & 11 & (\checkmark, \checkmark, \checkmark) \\  
\bottomrule  
\end{tabular} 
\end{table}
We concentrate on the TextWorld Cooking domain, which gained prominence through GATA~\citep{adhikari2020learning} and Microsoft's TextWorld Problems: A Language and Reinforcement Learning Challenge~\citep{trischler2019first}.  Following \cite{adhikari2020learning} and \cite{tuli2022learning}, we divide the games into five subsets with different difficulty levels. For easier games (with smaller difficulty levels), the recipe requires fewer ingredients, and the agent does not need to navigate through the world. For harder games, the agent is asked to navigate through different rooms to acquire various ingredients and cook the meal. The score is related to the relevant steps required in the cookbook of each game. If the agent completes a required step, it will earn one point. For the hardest game, the agent is required to finish 11 steps to complete this task. These steps involve opening certain containers to obtain the ingredients, cutting the ingredients as required (dice, slice, chop), and cooking the ingredient using the correct heat source (oven or stove). Statistics of the games are presented in Table \ref{tb:1}. In the dataset, expert demonstrations are provided in the form of walkthroughs and do not require human intervention for generation. Following previous work, we measure the performance of the algorithms using two metrics: normalized game points and game success rate. We test the game in 20 different games for each difficulty level. By averaging the points over the 20 games, and then dividing by the maximum score an agent can earn, we obtain the normalized game points. The game success rate is calculated as the percentage of games in which the agent successfully completes all the required steps.
We use GPT-4 as our base LLM.
\subsection {Few-shot performance}

We first test the few-shot performance of our method. We make a comparison between purely replay, tips summary using past history, and tips summary compared to expert trajectory. The latter two cases corresponds to our first and second scenario. For purely replay, we directly ingest the past trajectory as LLM's prompts to see if it can learn. Given that the input length is limited, we set the maximum number of trajectories as 3 in the experiments. For tips summary memory, the LLM agent is provided with the past failed actions and the tips generated from the last trial. We observe that with tips memory, the agent can collect more information using different trials and cumulatively achieve a high success rate (larger than 0.9) and high points earned (larger than 0.95) in the end, as shown in Figure \ref{fig:2}. With expert demonstration, the agent can learn faster and achieve zero error in all difficulty levels after at most 6 trajectories. This comparison shows that  concise tip summary assists the LLM in gathering information and making better decisions. With expert demonstrations, the LLM can learn more efficiently compared to solely relying on trial and error.

We further show an example about how tip summary can help the agent make better decision. In this log file (showing only relevant information), the agent is tasked with roasting a purple potato using a predefined ActionList in a text-based game. In the first attempt, the agent mistakenly uses the stove to cook the purple potato, resulting in frying it and losing the game. The agent then generates a tip to use the "roast" action. However, this is an invalid action, and it fails again. Next, the agent generates a second tip to use the "cook" command with the stove, suggesting a different heat setting or method to avoid frying. Unfortunately, in this text-based game, there are no commands to change the heat setting, leading to another failure. After the third attempt, the agent realizes it should try using an oven to roast the potato, and finally succeeds in the fourth trial. This demonstrates that improved tips can lead to successful learning.

\definecolor{trial1Color}{rgb}{0.980392, 0.858824, 0.776471}  
\definecolor{trial2Color}{rgb}{0.780392, 0.882353, 0.709804}  
\definecolor{trial3Color}{rgb}{1.0, 0.933333, 0.741176}  
\definecolor{trial4Color}{rgb}{0.909804, 0.85098, 0.952941}  

\begin{wrapfigure}{r}{0.5\textwidth}
\vspace{-3em}
\begin{mdframed}[backgroundcolor=trial1Color, linewidth=0pt] 
\textbf{First trial:} \\  
$>$agent: cook purple potato with stove \\  
system: You fried the purple potato. \\  
Tips to win the game next time: \\  
1. You should try roast the potato next time instead of cook purple potato with stove after dicing the purple potato;  
\end{mdframed}  
\begin{mdframed}[backgroundcolor=trial2Color, linewidth=0pt]  
\textbf{Second trial:} \\  
$>$agent: roast the potato  \\  
system: Invalid action. \\  
$>$agent:  cook purple potato with stove\\ 
system: You fried the purple potato. \\  
Tips to win the game next time: \\  
1. You should try  cook purple potato with stove  next time after you have dicing the purple potato, but make sure to use a different heat setting or method to avoid frying the purple potato;  
\end{mdframed}  
\begin{mdframed}[backgroundcolor=trial3Color, linewidth=0pt]  
\textbf{Third trial:} \\  
$>$agent: cook purple potato with stove \\  
system: You fried the purple potato. \\  
Tips to win the game next time: \\  
1. You should try cook purple potato with oven next time instead of cook purple potato with stove, as the recipe suggests roasting the purple potato rather than frying it;  
\end{mdframed}  
\begin{mdframed}[backgroundcolor=trial4Color, linewidth=0pt]  
\textbf{Forth trial:} \\  
$>$agent: cook purple potato with oven\\  
system: You roasted the purple potato. \\  
Your score has just gone up by one point.
\end{mdframed}  
\vspace{-2em}
\end{wrapfigure}

\definecolor{trial1Color}{rgb}{0.980392, 0.858824, 0.776471}  
\definecolor{trial2Color}{rgb}{0.780392, 0.882353, 0.709804}  
\definecolor{trial3Color}{rgb}{1.0, 0.933333, 0.741176}  
\definecolor{trial4Color}{rgb}{0.909804, 0.85098, 0.952941}  

\subsection{Zero-shot Performance}
In this experiment, we focus on training LLM agents using tips generated from the training set. We collect final tips, which are those leading to successful trials, from difficulty level 4 (the most challenging level) games in the training set. These tips are then sent to another LLM, which generates tips aimed at helping the agent become proficient across various game settings. The final tips can be found in the appendix. We used 10 games, which consist of 48 trajectories in total (averaging 4.8 trajectories per game) to generate the final tip.  We then test these tips on an unseen test set across different difficulty levels. To evaluate our LLM model with general tips as prompts, we compare it to state-of-the-art (SOTA) text-based game agents utilizing deep learning techniques, including TDQN~\citep{adhikari2020learning}, GATA~\citep{adhikari2020learning}, and ITL~\citep{tuli2022learning}. These models are trained on 100 different games for each difficulty level, with 100,000 episodes per level. Results for levels 0 to 3 are provided by the authors of~\citep{tuli2022learning}, while level 4 results are obtained from~\cite{adhikari2020learning}.

As shown in Figure 4, we prompt the LLM with general tips and test it on levels 1 to 4. Given level 0's simplicity, we do not provide tips for this level. Our findings reveal that the LLM with tips, using only 10 games and 48 trajectories, achieves performance comparable to SOTA methods for difficulty levels 0-3. For levels 3-4, LLM outperforms other methods,  due to the reasoning ability of the LLM that allows us to bypass the need to learn navigation and the correct tips generated in the same difficulty level, an aspect that hinders SOTA methods\citep{adhikari2020learning}.

\textbf{Limitation }Despite these achievements, the LLM agent underperforms in some lower difficulty level games. This underperformance can be attributed to two factors. First, the LLM fails to generate a general tip that addresses specific situations encountered in easier levels. For instance, in difficulty level 1, when an ingredient is already in its desired state according to the recipe (e.g., roasted or fried), the corresponding actions (roasting or frying) become unnecessary. By incorporating human-generated tips (see the appendix for the full list of tips), the LLM agent can achieve significantly higher points and success rates, as demonstrated in Table \ref{table:performance}. Second, the LLM's probabilistic nature leads to non-deterministic outputs, which occasionally cause the agent to disregard the tips and execute erroneous actions. Despite being provided with tips, the LLM agents will still make some mistakes due to their inherent probabilistic behavior. Nevertheless, even with this randomness, the LLM agent with general tips can outperform state-of-the-art (SOTA) agents specifically trained to excel in this task.

\begin{figure}[htb] 
    \centering
    \includegraphics[scale=0.44]{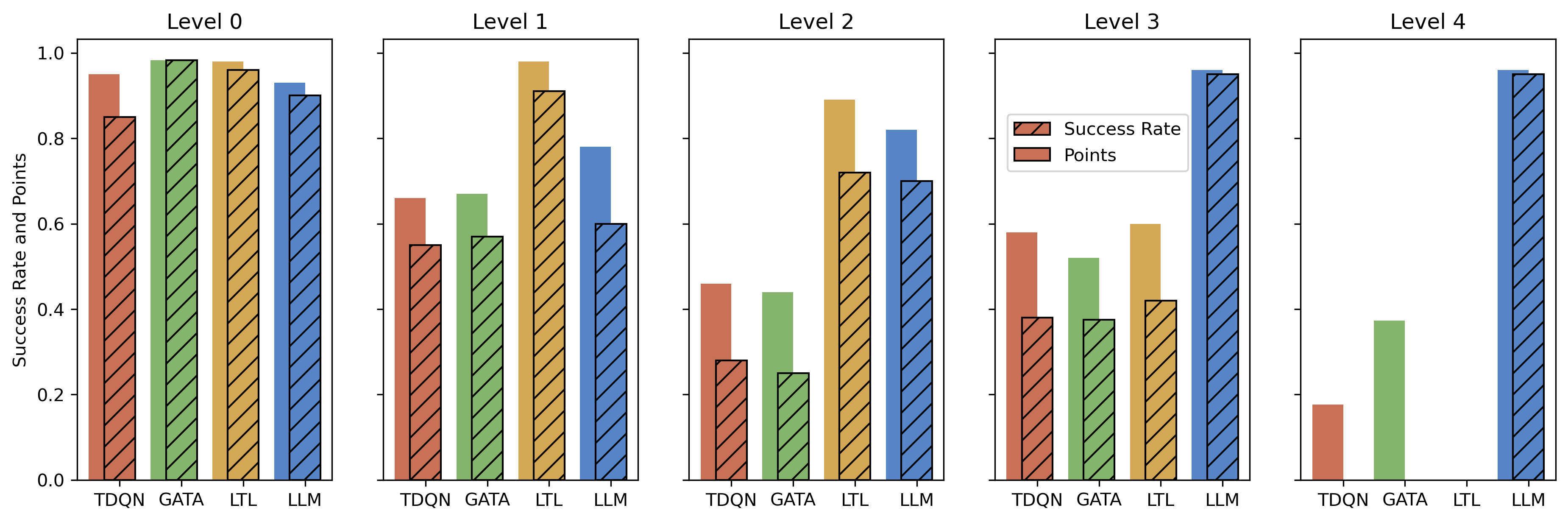}
    \caption{Performance of LLM as a text-based game agent compared to SOTA methods. Given that experimental results of difficulty level 4
     is not included in \cite{tuli2022learning}, we obtain data from \cite{adhikari2020learning} and some data is missing. }
    \label{fig:my_label}
\end{figure}
\begin{table}[ht]  
\caption{Performance of the LLM agent with human-generated tips across different difficulty levels.}  
\label{table:performance} 
\centering  
\footnotesize
\begin{tabular}{cccccccccc}  

\hline  
\multicolumn{2}{c }{Level 0} &\multicolumn{2}{c }{ Level 1}&\multicolumn{2}{c }{ Level 2}&\multicolumn{2}{c }{ Level 3}&\multicolumn{2}{c }{Level 4}\\
Points & Suc. Rate & Points  & Suc. Rate & Points  & Suc. Rate  & Points  & Suc. Rate  & Points  & Suc. Rate\\ \hline  
1 & 1 & 0.88 & 0.80 & 0.92 & 0.90 & 0.96 & 0.95 & 0.96 & 0.95 \\ \hline  
\end{tabular}   
\end{table}  

\vspace{-2em}
\section{Conclusion}
We introduce the novel concept of "Introspective Tips" as a powerful mechanism to improve the decision-making capabilities of LLM agents. Drawing inspiration from human introspection, this approach enables agents to extract and learn from generalized, high-level information that can be applied across various tasks and contexts. To effectively implement Introspective Tips, we propose a framework that dynamically adjusts prompts based on insights derived from past trajectories or expert demonstrations through introspection. This approach alleviates the burden of manual prompt crafting while empowering LLM agents with self-optimizing capabilities. By leveraging the rich common sense knowledge and generalization abilities of LLMs, our Introspective Tips paradigm outperforms SOTA methods in text-based game. Future work could focus on refining the framework for prompt generation, exploring more sophisticated methods for extracting tips from trajectories, and evaluating the effectiveness of introspective tips in a broader range of tasks and real-world applications.


\bibliography{main}
\newpage
\appendix
\section{Full Action List}
\begin{itemize}
\item \texttt{look()} \# describe the current room
\item \texttt{goal()} \# print the goal of this game
\item \texttt{inventory()} \# print player's inventory
\item \texttt{go(direction)} \# move the player north, east, south, or west
\item \texttt{examine(item)} \# examine something more closely
\item \texttt{eat(food)} \# eat edible food
\item \texttt{open(item)} \# open a door or a container
\item \texttt{close(item)} \# close a door or a container
\item \texttt{drop(item)} \# drop an item on the floor
\item \texttt{take(item)} \# take an item that is on the floor
\item \texttt{put(item, supporter)} \# place an item on a supporter
\item \texttt{take\_from(item, container)} \# take an item from a container or a supporter

\item \texttt{insert(item, container)} \# place an item into a container
\item \texttt{lock(item, key)} \# lock a door or a container with a key
\item \texttt{unlock(item, key)} \# unlock a door or a container with a key
\item \texttt{cook(food, heat\_source)} \# cook cookable food with something providing heat

\item \texttt{slice(food, sharp\_object)} \# slice cuttable food with something sharp
\item \texttt{chop(food, sharp\_object)} \# chop cuttable food with something sharp
\item \texttt{dice(food, sharp\_object)} \# dice cuttable food with something sharp

\item \texttt{prepare\_meal()} \# combine ingredients from inventory into a meal
\end{itemize}
\section{Gnenral Tips}
\subsection{ Tips generated using 10 games}
\begin{itemize}
\item 1. Always double-check the recipe before performing any actions, and follow the instructions closely.

\item 2. Make sure to gather all required ingredients and tools before starting to cook. Use the "examine(item)" and "take(item)" actions to interact with items in the environment.

\item 3. Pay attention to the cooking methods mentioned in the recipe, such as roasting or frying, and use the appropriate appliances like the oven and stove. To operate an appliance, use actions from the ActionList, such as "cook(food, heat\_source)".

\item 4. Use the correct actions from the ActionList to prepare the ingredients, such as "slice(food, sharp\_object)", "dice(food, sharp\_object)", or "chop(food, sharp\_object)".

\item 5. Keep track of your inventory and the state of each item to avoid making mistakes in the preparation process. Use the "inventory()" action to check your inventory.

\item 6. Follow the steps in the recipe in the correct order, ensuring each ingredient is prepared as instructed.

\item 7. Once all ingredients are prepared according to the recipe, use the "prepare\_meal()" action to combine them into a meal.

\item 8. If you encounter any difficulties or errors, refer back to the ActionList and the environment description to make sure you're using the correct actions and items.
\end{itemize}

\subsection{ Human-generated tips}
\begin{itemize}
\item 1 If you get stuck, use the actions goal(), examine(cookbook), inventory() and look() to figure out what you have and what you need to do.
\item 2 Before cooking an ingredient, always use inventory() first to check current state of the ingredient. 
If an ingredient is already in the desired state(roasted or fried) as per the recipe, there is no need to perform the actions (roast or fry) mentioned in the recipe.
\item 3 Try take(food) before you prepare(meal). After you take(food), use inventory() to check the status of the food. 
\item 4 Make sure the food is in your inventory before you prepare(meal). Try eat(meal) when the meal is prepared.  
\item 5 Dice, chop, and slice are different. 
\item 6 Fry and Roast are different. Use correct heat\_source to cook the ingredient. Oven is for roasting something, not frying something. 
\item 7 Use cook(food, oven) to roast something, instead of putting the food into the oven and wait.
\item8 Be careful when you cook(food, stove) or cook(food, oven),  make sure the food is not fried or roasted before. Otherwise you may burn the food and lose the game.
\end{itemize}

\section{Prompts }
Prompts for generating the tips:

\textbf{Generating tips for the first trial} 

Now you failed the game. Come up with tips for yourself to win the game next time.  Pay attention to steps that make you lose the game. Pay attention to those actions and come up with concise actions in the ActionList that you should try next time. Starting your response with  "Tips to win the game next time:"

\textbf{Generating tips for other trial} 

Now you failed the game. Come up with tips for yourself to win the game next time. 
Pay attention to steps that make you lose the game. Make a note on those actions and come up with concise actions in the ActionList that you should try next time. \textbf{Correct and improve the previous tips that I gave you if you think some of them are wrong.}

\textbf{Generating general tips } 

I want you to provide me all possible specific reasons that an agent may fail in a text-based game and summarize all tips for the agent to improve. I will present you several reasons and tips that  summarized by others before. In every file, the action that make an agent failed before and some tips to correct this error are provided.  Noticed that the tips provided may not be correct. For example, some actions suggested by the tips are not in the ActionList.
Correct this kind of error in the tips that you generate. This tips will be used by another rookie agent. So make sure to provide all necessary and userful information for the rookie agent. You should also include action in the ActionList that the agent should use to operate appliance.

\section{Sample Trajectory}
\setlength{\parindent}{0em}  
\noindent
Env:
You are hungry! Let's cook a delicious meal. Check the cookbook in the kitchen for the recipe. Once done, enjoy your meal!

{\centering
-= Bedroom =-\\}

\noindent 
You are in a bedroom. An ordinary kind of place.

You smell an intriguing smell, and follow it to a bed. Unfortunately, there isn't a thing on it. You move on, clearly infuriated by TextWorld.

 You don't like doors? Why not try going west, that entranceway is not blocked by one.

\textcolor{myColor}{$>$agent: go(west)}

\noindent Env: 

{\centering
-= Corridor =-\\}
\noindent You are in a corridor. An usual kind of place. I guess you better just go and list everything you see here.

There is an exit to the east. There is an exit to the north. There is an exit to the west. Don't worry, there is no door.

\textcolor{myColor}{ $>$agent
: go(north)}

Env: 

{\centering
-= Kitchen =-\\}
You find yourself in a kitchen. A standard kind of place.

You can see a closed fridge, which looks conventional, nearby. You can see a closed oven. What's that over there? It looks like it's a table. The table is massive. On the table you can make out a cookbook. You rest your hand against a wall, but you miss the wall and fall onto a counter! On the counter you make out a raw red potato, a red apple, a raw purple potato, a red hot pepper, a raw yellow potato, a banana and a knife. You see a stove. The stove is conventional. However, the stove, like an empty stove, has nothing on it.

There is a closed plain door leading north. You don't like doors? Why not try going south, that entranceway is not blocked by one. There is an exit to the west.

\textcolor{myColor}{$>$agent: examine(cookbook)}

Env: You open the copy of "Cooking: A Modern Approach (3rd Ed.)" and start reading:

Recipe \#1

---------

Gather all following ingredients and follow the directions to prepare this tasty meal.

Ingredients:
\setlength{\parindent}{2em}  

  red potato
  
  white onion
  
  yellow bell pepper
  
\setlength{\parindent}{0em}  

Directions:
\setlength{\parindent}{2em}  

  slice the red potato
  
  roast the red potato
  
  dice the white onion
  
  fry the white onion
  
  slice the yellow bell pepper
  
  fry the yellow bell pepper
  
  prepare meal

\setlength{\parindent}{0em}  
\textcolor{myColor}{$>$agent
: take(red\_potato)}

Env: You take the red potato from the counter.
Your score has just gone up by one point.

\textcolor{myColor}{$>$agent
: take(cutting\_tool)}

Env: You take the knife from the counter.

\textcolor{myColor}{$>$agent
: slice(red\_potato, cutting\_tool)}

Env: You slice the red potato.

Your score has just gone up by one point.

\textcolor{myColor}{$>$agent
: cook(red\_potato, oven)}

Env: You roasted the red potato.

Your score has just gone up by one point.

\textcolor{myColor}{$>$agent: go(south)}

Env: 

{\centering
-= Corridor =-\\}

You are in a corridor. An usual kind of place. I guess you better just go and list everything you see here.

There is an exit to the east. There is an exit to the north. There is an exit to the west. Don't worry, there is no door.

\textcolor{myColor}{$>$agent
: go(west)}

Env: 

{\centering
-= Bathroom =-\\}
You find yourself in a bathroom. An ordinary kind of place.

You make out a toilet. But oh no! there's nothing on this piece of garbage.

You need an exit without a door? You should try going east.

\textcolor{myColor}{$>$agent
: go(east)}

Env: 

{\centering
-= Corridor =-\\}
You are in a corridor. An usual kind of place. I guess you better just go and list everything you see here.

There is an exit to the east. There is an exit to the north. There is an exit to the west. Don't worry, there is no door.

\textcolor{myColor}{$>$agent
: go(east)}

Env: 

{\centering
-= Bedroom =-\\}

You are in a bedroom. An ordinary kind of place.

You smell an intriguing smell, and follow it to a bed. Unfortunately, there isn't a thing on it. You move on, clearly infuriated by TextWorld.

You don't like doors? Why not try going west, that entranceway is not blocked by one.

\textcolor{myColor}{$>$agent
: go(west)}

Env:

{\centering
-= Corridor =-\\}
You are in a corridor. An usual kind of place. I guess you better just go and list everything you see here.

There is an exit to the east. There is an exit to the north. There is an exit to the west. Don't worry, there is no door.

\textcolor{myColor}{$>$agent
: go(north)}

Env:

{\centering
-= kitchen =-\\}
You find yourself in a kitchen. A standard kind of place.

You can see a closed fridge, which looks conventional, nearby. You can see a closed oven. What's that over there? It looks like it's a table. The table is massive. On the table you can make out a cookbook. You rest your hand against a wall, but you miss the wall and fall onto a counter! On the counter you make out a red apple, a raw purple potato, a red hot pepper, a raw yellow potato and a banana. You see a stove. The stove is conventional. However, the stove, like an empty stove, has nothing on it.

There is a closed plain door leading north. You don't like doors? Why not try going south, that entranceway is not blocked by one. There is an exit to the west.

\textcolor{myColor}{$>$agent
: open(fridge)}

Env: You open the fridge, revealing a yellow bell pepper, a white onion, a carrot, a red onion, an orange bell pepper, a block of cheese and a raw pork chop.

\textcolor{myColor}{$>$agent
: take(white\_onion)}

Env: You take the white onion from the fridge.

Your score has just gone up by one point.

\textcolor{myColor}{$>$agent
: take(yellow\_bell\_pepper)}

Env: You take the yellow bell pepper from the fridge.

Your score has just gone up by one point.

\textcolor{myColor}{$>$agent: dice(white\_onion, cutting\_tool)}

Env: You dice the white onion.

Your score has just gone up by one point.

\textcolor{myColor}{$>$agent: cook(white\_onion, stove)}

Env: You fried the white onion.

Your score has just gone up by one point.

\textcolor{myColor}{$>$agent: slice(yellow\_bell\_pepper, cutting\_tool)}

Env: You slice the yellow bell pepper.

Your score has just gone up by one point.

\textcolor{myColor}{$>$agent: cook(yellow\_bell\_pepper, stove)}

Env: You fried the yellow bell pepper.

Your score has just gone up by one point.

\textcolor{myColor}{$>$agent: prepare\_meal()}

Env: Adding the meal to your inventory.

Your score has just gone up by one point.

\textcolor{myColor}{$>$agent: eat(meal)}

Env: You eat the meal. Not bad.
Your score has just gone up by one point.

{\centering
  *** The End ***\\}
                                                
You scored 11 out of a possible 11, in 23 turn(s).
\end{document}